\pdfoutput=1

\documentclass[runningheads]{llncs}

\usepackage[T1]{fontenc}
\usepackage{graphicx}
\usepackage{multirow}
\usepackage{amsmath,amssymb}
\usepackage{xcolor}
\usepackage{textcomp}
\usepackage{booktabs}
\usepackage{array}
\usepackage{url}

\usepackage{lineno}

\graphicspath{{figs/}}
\newif\ifanon
\anonfalse
\newcommand{\anon}[2]{\ifanon#1\else#2\fi}

\newcommand*\linenomathpatch[1]{%
  \expandafter\let\csname old#1\expandafter\endcsname\csname #1\endcsname
  \expandafter\let\csname oldend#1\expandafter\endcsname\csname end#1\endcsname
  \renewenvironment{#1}{\linenomath\csname old#1\endcsname}{\csname oldend#1\endcsname\endlinenomath}}
\newcommand*\linenomathpatchAMS[1]{\linenomathpatch{#1}\linenomathpatch{#1*}}
\AtBeginDocument{\linenomathpatchAMS{equation}\linenomathpatchAMS{align}}

\begin{document}

\title{NPU-Accelerated GNSS Signal Acquisition}
\titlerunning{\mbox{}} 

\ifanon
\author{Anonymous author(s)}
\institute{Affiliation withheld for anonymous review}
\else
\author{Taro Suzuki\thanks{Corresponding author}}
\institute{Department of Advanced Robotics, Chiba Institute of Technology, Narashino, Chiba, Japan\\
\email{taro.suzuki@chibatech.ac.jp}}
\fi
\authorrunning{\mbox{}} 

\maketitle

\begin{abstract}
On small single-board computers (SBCs), GNSS signal acquisition is a major CPU load. This paper proposes using the neural processing unit (NPU) of an SBC not for machine learning but as an accelerator for the parallel code search in GNSS signal acquisition. The discrete Fourier transform (DFT) used in acquisition is decomposed by the four-step DFT into matrix products of small DFTs and expressed by $1\times1$ convolutions, element-wise products, and transposes, which NPUs execute efficiently. Furthermore, we propose a method that divides the DFT, the non-coherent integration, and the peak detection between the CPU and the NPU according to the configuration of the NPU (an NPU with its own memory connected via PCIe, or an NPU sharing memory with the CPU). The proposed method is implemented in the open-source software GNSS receiver Pocket SDR and evaluated on a Raspberry Pi~5 combined with the external AX8850 NPU and on the NPU built into the SoC (RK3588) of the ROCK 5C, a small SBC of comparable size. Compared with acquisition by an optimized CPU implementation, the acquisition of short codes on the Raspberry Pi~5 with the AX8850 is 3.0--3.8 times faster. In the evaluation of the whole receiver, the number of tracking channels that reach their first lock within 30~s with the NPU is 1.5 times that with the CPU, and the time until 90\% of the channels are locked is reduced from 66~s to 26~s. Furthermore, because the CPU load of acquisition is reduced, the number of tracking channels that can be processed in real time increases from about 175 to about 230. These results show that using the NPU on SBCs with limited CPU resources is effective for achieving both GNSS signal acquisition and tracking. The implementation is publicly available as open source under the name PocketSDR-NPU.

\end{abstract}

\section{Introduction}\label{sec:intro}

A software GNSS receiver (GNSS SDR) performs the GNSS signal processing of IF data, digitized by an RF front-end, on a general-purpose computer~\cite{Pany2024GNSSSDR,Borre2007Software}. Because the GNSS signal processing is implemented in software, new signals and algorithms can be supported easily, and GNSS SDRs are widely used in research and education. Recent open-hardware RF front-ends can simultaneously receive signals on four or more frequencies from multiple constellations. For example, Pocket SDR~\cite{Takasu2026PocketSDR}, which is published as open hardware, provides four to eight RF front-end channels for the L1, L2, L5, and L6 bands. Receiver software such as GNSS-SDR~\cite{FernandezPrades2011GNSSSDR} and Pocket SDR is also available as open source.

There is a strong demand to run such receivers on small, low-power ARM single-board computers (SBCs) such as the Raspberry Pi. Receivers on SBCs would enable portable multi-frequency receivers and dense networks of receivers at low cost, and GNSS SDRs optimized for ARM processors have therefore been studied~\cite{TrogliaGamba2015eNGene,Miralles2022Neon}. However, the processing power of an SBC CPU is far below that of a general-purpose computer. Receiver processing consists of acquisition, which searches for the code phase and Doppler frequency of a signal, and tracking, which maintains the phase and frequency of an acquired signal. The tracking correlators impose a steady load proportional to the number of signals. In contrast, acquisition is a two-dimensional search over code phase and Doppler frequency, and a single search requires a large amount of computation even with the FFT-based parallel code search~\cite{VanNee1991Fast}. On SBCs, the acquisition load causes the following problems:
\begin{itemize}
\item Acquisition takes a long time, which delays the initial acquisition and lengthens the time from power-on to the first fix.
\item Reacquisition of signals lost due to blockage is delayed, reducing the number of usable signals in real environments such as urban areas.
\item Acquisition occupies the CPU and limits the CPU time available for tracking. If tracking cannot keep up with real time, the whole receiver fails.
\item For long codes such as L1C (10~ms) and L2CM (20~ms), a direct search over the full Doppler range requires a very large amount of computation, so the direct acquisition of long codes is impractical on SBCs.
\end{itemize}

Meanwhile, NPUs for machine-learning inference have become widely available on SBCs. For example, SoCs such as the Rockchip RK3588 include a built-in NPU, and NPUs that can be connected via PCIe (M.2), such as the AX8850, are available for the widely used Raspberry Pi~5. However, an NPU is an accelerator that executes matrix products and convolutions at low precision (e.g., INT8) at high speed for the inference of machine-learning models, and it is not used for ordinary SDR signal processing.

In this paper, we therefore propose using the NPU of an SBC as a dedicated processor for GNSS signal acquisition. We build a model that expresses the DFT of the parallel code search with operations that an NPU can execute ($1\times1$ convolution, element-wise product, and transpose), and offload the acquisition of the receiver to the NPU. The contributions of this paper are as follows:
\begin{itemize}
\item To the best of our knowledge, this is the first study to use the NPU of an SBC for conventional GNSS signal acquisition (the FFT-based parallel code search) rather than for machine learning.
\item We present a representation of the DFT suited to NPUs for executing the parallel code search and, according to the configuration of the NPU (its own memory with PCIe transfers, or memory shared with the CPU), a division of the DFT, the non-coherent integration, and the peak detection between the CPU and the NPU.
\item Compared with acquisition using an optimized CPU FFT, the search is 3.0--3.8 times faster on a Raspberry Pi~5 with an AX8850.
\item The direct search of long codes (L1CP and L2CM) over the full Doppler range becomes possible on SBCs within a practical time, and in the whole receiver, the initial acquisition becomes faster and the number of tracking channels that can be processed in real time increases from about 175 to about 230 (Raspberry Pi~5 with the AX8850).
\end{itemize}
The GNSS SDR with the NPU-accelerated acquisition is released as open source under the name PocketSDR-NPU\footnote{\anon{URL withheld for anonymous review}{\url{https://github.com/taroz/PocketSDR-NPU}}}.

\section{Related Work}\label{sec:related}

\subsection{Software GNSS Receivers}

The history and state of the art of GNSS SDRs are reviewed by Pany et al.~\cite{Pany2024GNSSSDR}. A textbook implementation is the receiver of Borre et al.~\cite{Borre2007Software}, and many software GNSS receivers have since been developed~\cite{Suzuki2014GNSSSDRLIB,Liaquat2024FGIGSRx}. Among them, GNSS-SDR~\cite{FernandezPrades2011GNSSSDR} and Pocket SDR~\cite{Takasu2026PocketSDR} are widely known as practical open-source receivers. Pocket SDR consists of open-hardware RF front-ends and a multi-frequency, multi-constellation GNSS SDR written in C, and it also runs on the Raspberry Pi. eNGene~\cite{TrogliaGamba2015eNGene} is a receiver running on an embedded ARM processor, and a signal processing library based on the ARM SIMD instructions (NEON) has been reported~\cite{Miralles2022Neon}. Nevertheless, even in GNSS SDRs optimized for ARM, signal acquisition remains one of the main computational loads on SBCs.

\subsection{Acceleration of Acquisition}

Acquisition is commonly based on the parallel code search, which obtains the correlations at all code phases at once by correlation in the frequency domain using the FFT~\cite{VanNee1991Fast,Akopian2005Fast,Borre2007Software}. Improvements to its speed~\cite{Leclere2014Modern} and the acceleration of correlation with SIMD instructions~\cite{Heckler2006SIMD,FernandezPrades2016Accelerating} have been reported. When the Doppler bin spacing is set equal to the FFT bin spacing, the FFT for another Doppler bin is obtained by circularly shifting the FFT of the received signal. Psiaki~\cite{Psiaki2004L2C} used this frequency shift for L2C acquisition, and Soltanian et al.~\cite{Soltanian2014DopplerFFT} estimated the Doppler from shifts in the FFT domain to reduce the number of IFFTs. In this paper, we refer to this technique as the shared DFT and use it as the base algorithm for both the CPU baseline and the NPU implementation.

Hardware such as GPUs and FPGAs has also been widely studied for accelerating acquisition. Pany et al.~\cite{Pany2010GPUAcq} ported acquisition to a GPU, and GPU-based real-time receivers have been developed~\cite{Hobiger2010GPU,Huang2013STARx}. A study of a receiver on an embedded GPU (Jetson TK1)~\cite{Park2016JetsonTK1} reported that the embedded GPU limits the real-time processing of the whole receiver. For FPGAs, acquisition architectures have been compared~\cite{Leclere2013Comparison} and high-sensitivity acquisition has been demonstrated~\cite{Majoral2024HighSensitivity}. Furthermore, dedicated integrated circuits (ASICs) for acquisition have been designed and developed, including a VLSI acquisition engine for multiple constellations and frequencies~\cite{Yang2018VLSIAcq}, an acquisition accelerator balancing energy efficiency and accuracy~\cite{Jeong2025ISSCC}, a mixed-signal accelerator with charge-based correlation~\cite{Li2025MatchedFilterJSSC}, and an acquisition macro using SRAM processing-in-memory~\cite{Gan2026SRAMPIM}. The continued development of dedicated circuits indicates that the computational load and energy consumption of acquisition are important issues for receivers. However, these methods require GPUs, FPGAs, or dedicated ASICs, unlike the approach of this study, which uses an NPU built into or added to a commercial SBC.

\subsection{Acquisition of Long Codes}

For signals with long codes, such as GPS L2CM (20~ms) and L1C (10~ms), the number of code phases to search grows with the code length, and the number of Doppler bins also grows because the bin spacing becomes narrower. Therefore, ordinary receivers narrow the search space by assisted acquisition, using the time and Doppler of a previously acquired signal with a short code such as L1 C/A~\cite{VanDiggelen2009AGPS}. Efficient acquisition methods for the long codes of L2C~\cite{Psiaki2004L2C,Qaisar2012L2C} and L1C~\cite{MacchiGernot2010L1C} have also been proposed. Direct acquisition without assistance has the advantage of acquiring signals without relying on other signals. Direct acquisition of the long P code by a large-scale FFT~\cite{Li2008DirectPCodeFFT} and fast direct acquisition of L2CL~\cite{Ziedan2010L2CLDirect} have been proposed, but their computational and memory requirements are large, which makes them impractical on the low-performance CPUs of SBCs.

\subsection{Signal Processing on AI Accelerators}

AI accelerators such as NPUs and TPUs have been applied to signal processing beyond machine learning, for example, large-scale DFTs computed as matrix products on a TPU~\cite{Lu2021LargeScaleDFT}, DFT/FFT algorithms for NPUs~\cite{Li2024NPUFourier,Yao2026DRFFT}, and a radio-astronomy correlator on GPU Tensor Cores~\cite{Romein2021TensorCore}. Yao et al.~\cite{Yao2026DRFFT} showed that the FFT must be restructured to fit the matrix units and memory hierarchy of an NPU. In GNSS, an autoencoder that compresses IF data has been run on an Edge TPU~\cite{Wegner2025EdgeTPU}, but acquisition and correlation were not computed on the TPU. Deep-learning-based acquisition~\cite{BorhaniDarian2023DL,Moradi2023CNN} replaces the detection in the correlation results (the cross-ambiguity function) with a neural network, whereas the correlation itself is still computed on the CPU.

To the best of our knowledge, no previous study has executed the FFT-based parallel code search of GNSS signals itself on the NPU of an SBC. The proposed method differs from previous work in that it accelerates signal acquisition, without building special hardware, by using an NPU that is already on an SBC or can be added at low cost, for GNSS signal acquisition instead of machine-learning inference.

\section{Method}\label{sec:method}

\subsection{System Overview and NPU Configurations}

\begin{figure}[t]
\centering
\includegraphics[width=\linewidth]{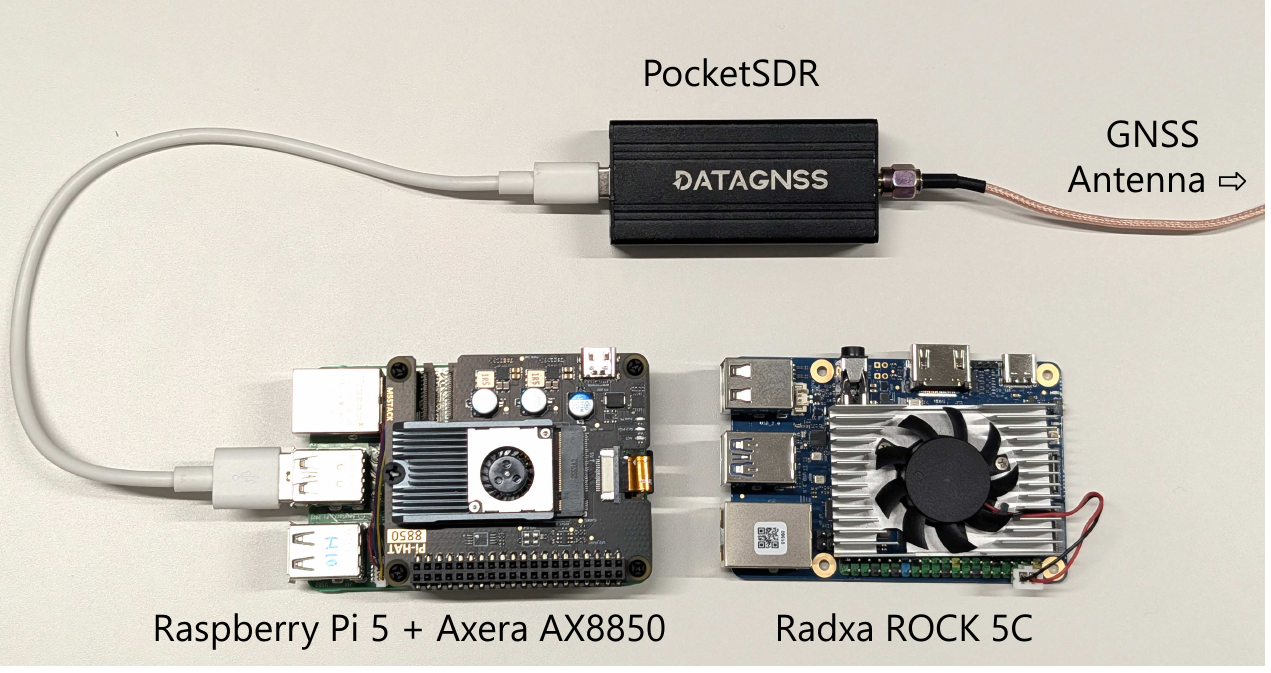}
\caption{SBCs and NPUs used in this study. The Raspberry Pi~5 with the AX8850 and the Radxa ROCK 5C are used, with the Pocket SDR FE as the GNSS front-end}\label{fig:photo}
\end{figure}

NPUs available for SBCs can be divided into two configurations according to how they are connected to the host CPU: (i) an NPU with its own memory connected to the host via PCIe or a similar interface, and (ii) an NPU built into the SoC that shares memory with the CPU. In configuration (i), the inputs and outputs are transferred to and from the host, so the amount and number of transfers determine the performance. In configuration (ii), the overhead of passing the inputs and outputs is small, and the computation on the NPU is the bottleneck. To evaluate NPUs of the two configurations, we target the following two combinations, which differ in the connection to the host (Table~\ref{tab:sbc}, Fig.~\ref{fig:photo}):
\begin{itemize}
\item \textbf{Dedicated-memory type: Raspberry Pi~5 with AX8850}: The AX8850 is an external NPU on an M.2 card (M5Stack LLM-8850) connected to the Raspberry Pi~5 via PCIe (Gen2 x1). The card has its own DDR memory (4~GB), and the host application can allocate the NPU memory and specify it directly as the inputs and outputs of an inference. Models are converted to the axmodel format with the model conversion tool Pulsar2 and executed with the AXCL runtime.
\item \textbf{Shared-memory type: ROCK 5C (RK3588)}: The ROCK 5C is an SBC of the same size as the Raspberry Pi~5. Its SoC, the RK3588, includes four Cortex-A76 and four Cortex-A55 CPU cores and a three-core NPU, and the NPU shares memory with the CPU. Models are converted to the RKNN format with the model conversion tool RKNN-Toolkit2 and executed on the NPU.
\end{itemize}

\begin{table}[t]
\caption{Specifications of the SBCs and NPUs}\label{tab:sbc}
\centering\small
\begin{tabular}{@{}>{\raggedright\arraybackslash}p{0.16\linewidth}>{\raggedright\arraybackslash}p{0.39\linewidth}>{\raggedright\arraybackslash}p{0.39\linewidth}@{}}
\toprule
 & Raspberry Pi 5 + AX8850 & ROCK 5C (RK3588) \\
\midrule
CPU & Cortex-A76 $\times$4, 2.4 GHz & Cortex-A76 $\times$4, 2.4 GHz + Cortex-A55 $\times$4, 1.8 GHz \\
Memory & 8 GB & 8 GB \\
NPU & AX8850 (24 TOPS INT8, 3 cores, M5Stack LLM-8850, 4 GB DDR on the card, PCIe Gen2 x1) & RK3588 built-in NPU (6 TOPS, 3 cores) \\
NPU runtime (model conversion tool) & AXCL 3.6.5 (Pulsar2 8.0) & RKNN runtime 2.3.0 (RKNN-Toolkit2 2.3) \\
OS & Debian 13 (kernel 6.18) & Radxa Debian 12 (kernel 6.1) \\
\bottomrule
\end{tabular}
\end{table}

\begin{figure}[t]
\centering
\includegraphics[width=\linewidth]{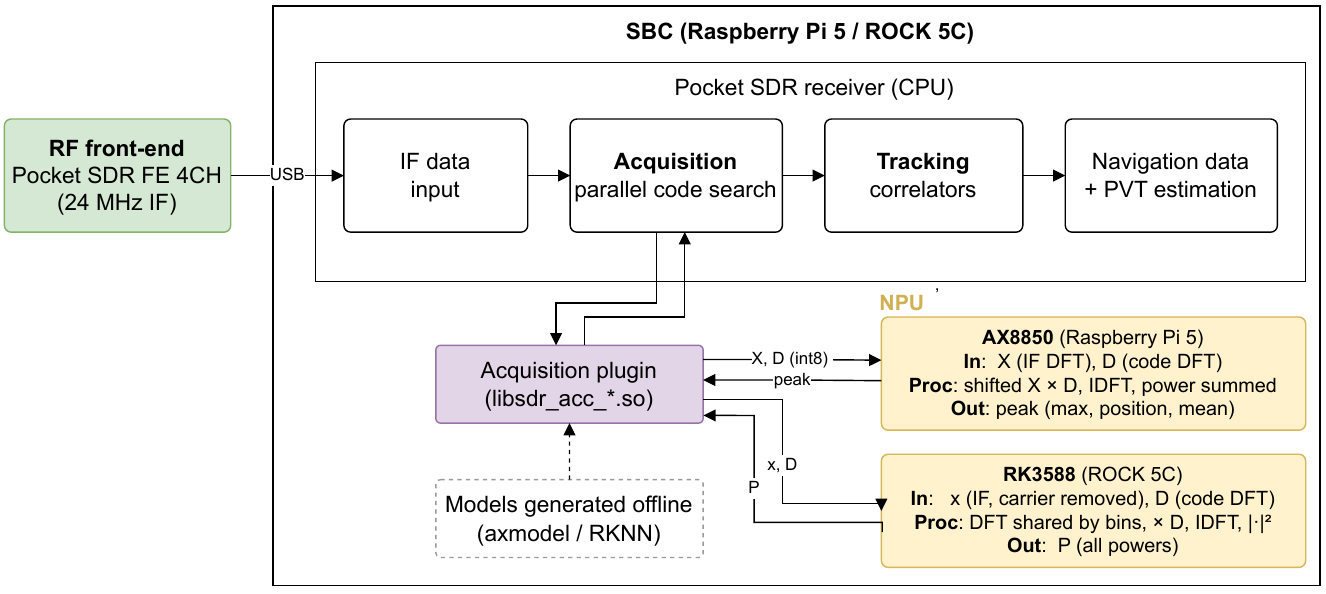}
\caption{Structure of the GNSS signal processing. A plugin executes acquisition on the NPU, running the parallel code search with inference models built in advance for each NPU}\label{fig:arch}
\end{figure}

We use Pocket SDR~\cite{Takasu2026PocketSDR} as the RF front-end and the base software of the SDR, and integrate the NPU-based acquisition into it. The RF front-end of Pocket SDR (Pocket SDR FE) has two, four, or eight RF channels, each receiving either the L1 band or the L2/L5/L6 bands, with ADC sampling frequencies of up to 48~MHz. In this study, we use the four-channel Pocket SDR FE shown in Fig.~\ref{fig:photo} with a sampling frequency of 24~MHz. The receiver software is written in C and supports the acquisition, tracking, navigation data decoding, and positioning of almost all signals of GPS, GLONASS, Galileo, QZSS, BeiDou, NavIC, and SBAS. Signals are processed in one thread per channel, and acquisition uses the FFT-based parallel code search. In this study, only the parallel code search of acquisition is executed on the NPU, without changing the structure of the receiver. Figure~\ref{fig:arch} shows the structure of the GNSS signal processing. We build a plugin that executes acquisition on the NPU and realize NPU-based acquisition by running the parallel code search with inference models built in advance for each NPU.

\subsection{Base Acquisition Method}\label{sec:shared}

This section describes the base acquisition method common to the CPU and the NPU. Acquisition in Pocket SDR is a circular correlation in the frequency domain between IF data spanning two code periods $T$ and a code replica of one period padded with zeros to the same length. Let $f_s$ be the sampling frequency. The IF data are divided into correlation segments of two code periods ($M = 2 T f_s$ samples) shifted by one code period $T$; let $x_w[n]$ ($n = 0, \ldots, M-1$) be the IF data of the $w$-th segment and $c[n]$ the zero-padded code replica. With the carrier frequency $f_k$ of Doppler bin $k$ (the IF center frequency plus the Doppler frequency), the DFT of the IF data after the carrier wipe-off and the correlation at code phase $\tau$ are
\begin{equation}
X_{k,w}[m] = \sum_{n=0}^{M-1} x_w[n]\, e^{-j 2\pi f_k n / f_s}\, e^{-j 2\pi m n / M},
\label{eq:dft}
\end{equation}
\begin{equation}
C_{k,w}[\tau] = \frac{1}{M} \sum_{m=0}^{M-1} X_{k,w}[m]\, D^{*}[m]\, e^{j 2\pi m \tau / M}, \quad \tau = 0, \ldots, M/2 - 1,
\label{eq:corr}
\end{equation}
where $D[m]$ is the DFT of $c[n]$, computed once per signal. Because the second half of the code replica is zero, the correlations for $\tau < M/2$ are not affected by the circular wrap-around with the second half of the IF data in the segment. The correlation powers of the $N_W = T_{\mathrm{acq}} / T$ segments within the non-coherent integration time $T_{\mathrm{acq}}$ (20~ms) are summed (20 for L1 C/A, 5 for E1B, 2 for L1CP, and 1 for L2CM), which is the non-coherent integration:
\begin{equation}
P_k[\tau] = \sum_{w=0}^{N_W - 1} \left| C_{k,w}[\tau] \right|^2 .
\label{eq:ncoh}
\end{equation}
The Doppler bins cover $\pm f_{\max}$ ($f_{\max} = 5$~kHz) with a spacing of $\Delta f = 0.5 / T$:
\begin{equation}
f_k = f_{\mathrm{IF}} - f_{\max} + k\, \Delta f, \quad k = 0, \ldots, K-1, \quad K = 2 f_{\max} / \Delta f + 1 .
\label{eq:bins}
\end{equation}
The number of Doppler bins $K$ is 21 for signals with a code period of 1~ms, 81 for E1B (4~ms), 201 for L1CP (10~ms), and 401 for L2CM (20~ms) (Table~\ref{tab:size}). The code phase and the Doppler frequency are obtained from the maximum $P_{\max}$ of $P_k[\tau]$ and its position, and the $C/N_0$ is estimated with the mean $\bar{P}$ of all powers as

\begin{table}[t]
\caption{Signals and search sizes (Doppler bins over $\pm$5 kHz with a spacing of $0.5/T$)}\label{tab:size}
\centering
\begin{tabular}{@{}llll@{}}
\toprule
Signal ($f_s$) & Code period $T$ & DFT size $M$ & Doppler bins (spacing) \\
\midrule
L1CA, B1I (8 MHz) & 1 ms & 16000 & 21 (500 Hz) \\
L5I (24 MHz) & 1 ms & 48000 & 21 (500 Hz) \\
E1B (8 MHz) & 4 ms & 64000 & 81 (125 Hz) \\
L1CP (8 MHz, direct) & 10 ms & 160000 & 201 (50 Hz) \\
L2CM (8 MHz, direct) & 20 ms & 320000 & 401 (25 Hz) \\
\bottomrule
\end{tabular}
\end{table}

\begin{equation}
\widehat{C/N_0} = 10 \log_{10} \frac{P_{\max} - \bar{P}}{\bar{P}\, T} \quad [\mathrm{dB\text{-}Hz}].
\label{eq:cn0}
\end{equation}
A signal is detected if the estimate exceeds a threshold (34~dB-Hz).

\textbf{Shared DFT}: Computing Eqs.~(\ref{eq:dft}) and (\ref{eq:corr}) directly for each bin requires $K$ DFTs and $K$ IDFTs per segment. Here, the bin spacing $\Delta f = 0.5/T$ equals the DFT bin spacing $f_s / M = f_s / (2 T f_s) = 0.5/T$. Substituting $f_k = f_0 + k\, f_s / M$ into Eq.~(\ref{eq:dft}) gives
\begin{equation}
X_{k,w}[m] = \sum_{n=0}^{M-1} x_w[n]\, e^{-j 2\pi f_0 n / f_s}\, e^{-j 2\pi (m + k) n / M} = X_{0,w}[(m + k) \bmod M],
\label{eq:share}
\end{equation}
that is, the DFT for bin $k$ equals the DFT $X_{0,w}$ for the first bin ($k = 0$) circularly shifted by $k$. With this property~\cite{Psiaki2004L2C,Soltanian2014DopplerFFT}, the DFT of the IF data needs to be computed only once per segment, and for each bin only the IDFT of the product of the shifted $X_{0,w}$ and the code DFT is computed. We refer to this as the shared DFT. The number of FFTs per search is reduced from $2 K N_W$ to $(K + 1) N_W$, which approaches one half as the number of bins increases.

\textbf{Decimation}: The Pocket SDR FE records IF data at a high sampling frequency (24~MHz) to receive wideband signals such as L5. Processing narrowband signals such as GPS L1 C/A and Galileo E1 at this rate, however, is wasteful. PocketSDR-NPU therefore extracts the sub-band of an RF channel containing narrowband signals with a polyphase FIR filter ($12D+1$ taps) and decimates it by a factor of $D$ before acquisition and tracking. In this study, the narrowband signals in the L1 band and L2C are decimated to 8~MHz ($D = 3$), whereas the wideband signals L5/E5a/E5b/B2a are processed at 24~MHz. The decimation reduces the DFT size $M$ by a factor of three and the FFT computation by slightly more than that.

The shared DFT and the decimation are common algorithms applied to both the CPU and the NPU.

\subsection{Representation of the Parallel Code Search on the NPU}\label{sec:npu}

\subsubsection{Decomposition of the DFT by the four-step DFT.}

An NPU can execute convolutions, matrix products, element-wise operations, and data rearrangements, but it has no dedicated FFT unit. Computing the DFT directly as a product with the $M \times M$ DFT matrix would require $M^2$ complex multiply-accumulates per DFT and far too many weights ($2.6 \times 10^8$ elements for $M = 16000$). We therefore decompose the DFT into small matrix products using the four-step DFT of Bailey~\cite{Bailey1990FourStep}. With $M = A \times B$, the indices split as $n = B n_1 + n_2$ ($0 \le n_1 < A$, $0 \le n_2 < B$) and $m = m_1 + A m_2$ ($0 \le m_1 < A$, $0 \le m_2 < B$), and $W_N = e^{-j 2\pi / N}$,
\begin{equation}
X[m_1 + A m_2] = \sum_{n_2=0}^{B-1} W_B^{n_2 m_2} \left[ W_M^{n_2 m_1} \left( \sum_{n_1=0}^{A-1} W_A^{n_1 m_1}\, x[B n_1 + n_2] \right) \right].
\label{eq:fourstep}
\end{equation}
The $M$-point DFT thus consists of four steps: (1) $B$ DFTs of length $A$ (matrix $F_A$), (2) element-wise multiplication by the twiddle factors $W_M^{n_2 m_1}$, (3) a transpose, and (4) $A$ DFTs of length $B$ (matrix $F_B$). Arranging the IF data in an $A \times B$ matrix $\mathbf{x}$ with $[\mathbf{x}]_{n_1 n_2} = x[B n_1 + n_2]$, this can be written in matrix form as
\begin{equation}
\mathbf{X} = \left( \left( F_A\, \mathbf{x} \right) \odot \mathbf{T} \right) F_B, \quad [\mathbf{T}]_{m_1 n_2} = W_M^{n_2 m_1},
\label{eq:matrix}
\end{equation}
where $\odot$ denotes the element-wise product and $[\mathbf{X}]_{m_1 m_2} = X[m_1 + A m_2]$. The number of complex multiply-accumulates is $M (A + B)$, about 1/63 of that of the direct DFT for $M = 16000$ ($A = 125$, $B = 128$). This is more than that of the FFT (of the order of $M \log_2 M$), but the structure suits the matrix units of the NPU and can be computed efficiently.

On the NPU, steps (1) and (4) are $1\times1$ convolutions along the channel axis with the DFT matrices as weights, step (2) is an element-wise product with constants, and step (3) is a transpose. The real and imaginary parts of complex values are held in separate channels, and the product of a complex weight $W = W_r + j W_i$ and an input $Z = Z_r + j Z_i$ is computed with three real convolutions:
\begin{equation}
\begin{split}
\mathrm{Re}(W Z) &= W_r (Z_r + Z_i) - (W_r + W_i) Z_i, \\
\mathrm{Im}(W Z) &= W_r (Z_r + Z_i) + (W_i - W_r) Z_r ,
\end{split}
\label{eq:cplx}
\end{equation}
The IDFT has the same structure, and its last stage outputs only the $R$ rows that contain the required code phases ($M/2$), where $R B \ge M/2$.

\subsubsection{Chunks and where the DFT is computed.}

\begin{figure}[t]
\centering
\includegraphics[width=\linewidth]{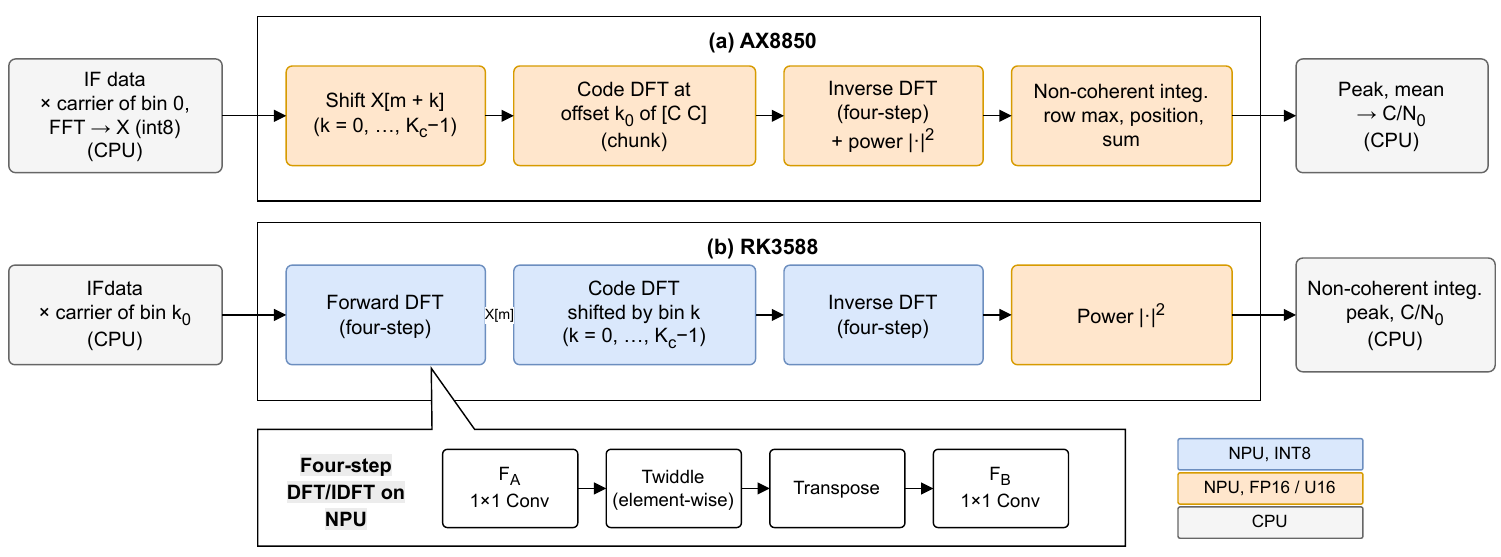}
\caption{Four-step DFT and shared DFT on the NPU. (a) Dedicated-memory type (AX8850): the forward DFT is computed only once by the FFT on the CPU, and the non-coherent integration and the row-wise maximum, position, and sum are computed on the NPU. (b) Shared-memory type (RK3588): the forward DFT is also computed on the NPU}\label{fig:dft}
\end{figure}

One inference processes a chunk of $K_c$ Doppler bins. Let $k_0$ be the first bin of a chunk. Using the shared DFT of Eq.~(\ref{eq:share}), the correlation for bin $k_0 + k$ ($0 \le k < K_c$) in the chunk can be written as
\begin{equation}
\begin{split}
\left| \mathrm{IDFT}\left[ X_{0}[m + k_0 + k]\, D^{*}[m] \right][\tau] \right|
&= \left| \mathrm{IDFT}\left[ X_{0}[m + k]\, D^{*}[m - k_0] \right][\tau] \right| \\
&= \left| \mathrm{IDFT}\left[ X_{0}[m]\, D^{*}[m - k_0 - k] \right][\tau] \right|,
\end{split}
\label{eq:shift}
\end{equation}
where the indices are taken modulo $M$. Eq.~(\ref{eq:shift}) holds because substituting $m' = m + k$ only multiplies the result of the IDFT by a phase factor $e^{-j 2\pi k \tau / M}$ of unit magnitude. In other words, a shift of the bin can be expressed by shifting either the DFT of the IF data or the code DFT. We use this freedom to choose where the DFT of the IF data is computed according to the configuration of the NPU. Figure~\ref{fig:dft} shows the processing flows of the configuration in which the CPU computes the DFT of the IF data ((a) dedicated-memory type, AX8850) and the configuration in which the NPU computes it ((b) shared-memory type, RK3588).
\begin{itemize}
\item \textbf{Dedicated-memory type (AX8850)}: The forward DFT is computed on the CPU. For each correlation segment, the CPU computes the DFT $X_{0,w}$ of the IF data after the wipe-off of the carrier of the first bin (frequency $f_0$) only once with an FFT and passes it to the NPU as 8-bit integers. The model forms the $K_c$ bins of the chunk as the shifts $X_0[m + k]$ and computes their products with the code DFT and the IDFTs. The chunk offset $k_0$ is expressed by the shifted code DFT $D^{*}[m - k_0]$. The IF data are input once per segment and the code DFT once per signal, which keeps the transfers to and from the host small.
\item \textbf{Shared-memory type (RK3588)}: The forward DFT is computed on the NPU. The CPU multiplies the IF data by the carrier of bin $k_0$, $e^{-j 2\pi f_{k_0} n / f_s}$, and passes the result to the NPU. The model computes the DFT of the IF data once and shares it among the $K_c$ bins of the chunk; the bins differ only in the product with the shifted code DFT $D^{*}[m - k]$ (an input that the CPU prepares for each signal) and the IDFT. The CPU only performs the carrier wipe-off, and most of the computation on the NPU is the IDFT of each bin.
\end{itemize}
In both cases, the number of inferences per search is $\lceil K / K_c \rceil N_W$, and the output is the correlation power $|C_{k,w}[\tau]|^2 / 64$, where the factor 1/64 scales the output down to keep it within the numerical range of the NPU. All correlation segments of the non-coherent integration (20 for L1 C/A) are passed in a single plugin call, and the plugin runs the NPU inferences for each segment and chunk internally. This eliminates the latency of one call per segment and allows the NPU to run the inferences back to back.

\subsubsection{Non-coherent integration and peak detection.}

The non-coherent integration of Eq.~(\ref{eq:ncoh}) and the computation of the maximum and mean in Eq.~(\ref{eq:cn0}) involve the powers at all code phases and bins ($401 \times 160000$ values for L2CM). On the RK3588, which shares memory with the CPU, the CPU reads the outputs of the NPU directly and computes the non-coherent integration and the maximum and mean.

On the AX8850, which is attached via PCIe, transferring all powers to the host memory would dominate the search time (about 67~MB for L1CP). The AX8850 therefore performs the non-coherent integration and the peak detection on the NPU. Each inference processes $n_w$ correlation segments, and the model has an additional input $P^{(a)}$ for the sum of the powers of the preceding inferences, so that the $j$-th inference outputs
\begin{equation}
P_k^{(j)}[\tau] = P_k^{(j-1)}[\tau] + \sum_{i=0}^{n_w-1} \left| C_{k, j n_w + i}[\tau] \right|^2 .
\label{eq:acc}
\end{equation}
The output stays in the NPU memory and is passed as the input $P^{(a)}$ of the next inference, so the non-coherent integration causes no transfer to or from the host. The last inference additionally arranges the code phases $\tau$ in $R$ rows $\times$ $B$ columns ($\tau = B p_1 + p_2$, $R B = M/2$) and outputs the maximum, its position, and the sum for each row $p_1$:
\begin{equation}
\begin{split}
P^{\max}_{k}[p_1] &= \max_{p_2} P_k[B p_1 + p_2], \quad P^{\arg}_{k}[p_1] = \mathop{\mathrm{arg\,max}}_{p_2} P_k[B p_1 + p_2], \\
P^{\mathrm{sum}}_{k}[p_1] &= \sum_{p_2} P_k[B p_1 + p_2].
\end{split}
\label{eq:red}
\end{equation}
The CPU receives only these $3 K R$ values to obtain $P_{\max}$, its position, and $\bar{P}$ in Eq.~(\ref{eq:cn0}) ($M = A \times B$ is factorized so that $R B = M/2$). The powers of the two neighboring bins used for the fine Doppler estimation are read as two values from $P_k[\tau]$ in the NPU memory. The output transfer per search is thus reduced from 67~MB to about 0.5~MB for L1CP.

\subsubsection{Quantization.}

To exploit the performance of the NPU, the model is quantized to integers. The quantization of an NPU represents a real value $r$ by an integer $q = \mathrm{round}(r / s) + z$ with a scale $s$ and a zero point $z$~\cite{Jacob2018Quantization}, and the scales are determined for each layer from the range of the calibration data. To match the input ranges to those of the calibration, the IF data (or their DFT on the AX8850) are normalized to unit RMS and the code DFT is normalized by its maximum magnitude:
\begin{equation}
\tilde{x}[n] = \frac{x[n]}{\sqrt{\frac{1}{M} \sum_{n} |x[n]|^2}}, \quad \tilde{D}[m] = \frac{D[m]}{\max_{m} |D[m]|} .
\label{eq:norm}
\end{equation}
The normalization factors are restored to the output powers so that they have the same scale as the CPU results. The precision follows the formats supported by each NPU. On the AX8850, the inputs are INT8, all internal operations are 16-bit integers (U16), and the non-coherent integration of Eq.~(\ref{eq:acc}) and the sums of Eq.~(\ref{eq:red}) are FP32. On the RK3588, the main operations are INT8, and the last IDFT stage and the power computation are kept in FP16. If the power were also INT8, the noise power would be quantized to only a few levels and the correlation peaks of strong signals would saturate. The quantization is calibrated with synthetic signals of the target codes ($C/N_0$ of 30--60~dB-Hz). The calibration data always include a signal with the maximum $C/N_0$ so that the calibrated range covers the strongest signals.

\subsection{Software Structure}\label{sec:sw}

PocketSDR-NPU keeps the structure of the receiver (channels, tracking, navigation data, and positioning) as in Pocket SDR (version 0.21) and delegates only the parallel code search to a plugin loaded at runtime (libsdr\_acc\_<name>.so) (Fig.~\ref{fig:arch}). The plugin interface consists only of simple tensors---the IF data (complex INT8, $M$ samples $\times$ $N_W$ correlation segments), the code DFT $D^{*}[m]$, and the Doppler bin frequencies $f_k$ as inputs and the correlation powers $P_k[\tau]$ as the output---and does not depend on the data structures of the receiver. In addition, an optional function returns only the peak values (the maximum, its position, the mean, and the powers of the neighboring bins) instead of the correlation powers. If the plugin supports this function, the receiver uses it to obtain the $C/N_0$ of Eq.~(\ref{eq:cn0}) and the Doppler frequency. Tracking runs on the CPU, and for real-time processing on SBCs, we added the assignment of CPU cores to acquisition and tracking, a real-time priority for the IF data input thread, and a function that suspends new searches while tracking is delayed. The source code of PocketSDR-NPU, the model generation scripts, the generated models, and the evaluation scripts are available on GitHub.

\section{Evaluation Conditions}\label{sec:cond}

To evaluate the proposed NPU-based acquisition, IF data of real GNSS signals were recorded with the RF front-end. The antenna was installed on the roof of a building with an open sky view, and a Pocket SDR FE 4CH recorded four RF channels (L1, L2, L5, and E5b) at a sampling frequency of 24~MHz. The first 300~s of the IF data recorded on October 9, 2026 were used, in which about 189 signals were receivable. As described in Section~\ref{sec:shared}, the narrowband signals in the L1 band and L2C were decimated to 8~MHz, whereas the wideband signals L5/E5a/E5b/B2a were processed at 24~MHz. Table~\ref{tab:size} lists the search size for each signal.

For the CPU search used for comparison, two FFT libraries were used: PocketFFT~\cite{PocketFFT}, the default of Pocket SDR, and the optimized FFTW~\cite{Frigo2005FFTW3}, with wisdom generated in advance for the DFT sizes used in the evaluation. The hardware is listed in Table~\ref{tab:sbc}. The CPU frequency governor was set to performance, and on the ROCK 5C the DDR and NPU frequency governors were also set to performance.

\section{Evaluation 1: Representations of the DFT on the NPU}\label{sec:evaldft}

The DFT by matrix products proposed for the NPU computes the small DFTs as matrix products and therefore requires more operations than the ordinary FFT. We therefore compared the proposed DFT by matrix products with an NPU implementation of the ordinary FFT algorithm. The DFT size was $M = 16000$, the same as for L1 C/A (8~MHz), and one inference computed 21 DFTs (the number of Doppler bins of L1 C/A). The following two models were compared:
\begin{itemize}
\item \textbf{DFT by matrix products (proposed)}: With $M = 125 \times 128$, the two DFTs of 125 and 128 points are each computed by $1\times1$ convolutions (matrix products).
\item \textbf{Radix FFT}: With $M = 5^3 \times 2^7$, the FFT consists of three radix-5 stages (5-point DFTs computed by $1\times1$ convolutions) and seven radix-2 stages (butterflies of additions and subtractions).
\end{itemize}
Both models were quantized to 16 bits in the same way, and the execution time of one inference on the NPU (AX8850) was measured as the mean of 200 inferences. Table~\ref{tab:dftrep} shows the number of complex multiply-accumulates and the execution time of each method.

\begin{table}[t]
\caption{Representations of the DFT on the NPU (AX8850, $M = 16000$)}\label{tab:dftrep}
\centering
\begin{tabular}{@{}lccc@{}}
\toprule
Method & Stages & Complex multiply-accumulates & Execution time \\
\midrule
DFT by matrix products & 2 & $85.3 \times 10^6$ & 0.88 ms \\
Radix FFT & 10 & $10.4 \times 10^6$ & 3.08 ms \\
\bottomrule
\end{tabular}
\end{table}

Although the number of complex multiply-accumulates of the NPU implementation of the ordinary FFT algorithm was about 1/8 of that of the proposed method, its execution time was about 3.5 times longer. The matrix units of the NPU execute large matrix products efficiently, whereas in the ordinary FFT algorithm most of the computation consists of small operations and of the transposes and element-wise products of each stage, which cannot exploit the matrix units. This result confirms that the proposed DFT by matrix products is a computation method suited to NPUs.

\section{Evaluation 2: Acquisition of Short Codes}\label{sec:eval1}

\subsection{Acquisition Time}

For L1 C/A, E1B, B1I (8~MHz), and L5I (24~MHz), all PRN codes were searched under the same conditions (non-coherent integration of 20~ms and a Doppler range of $\pm$5~kHz), and the search time and CPU time (time processed by the CPU) per signal were measured. The search was called from a single thread, and the times are averages over five trials. On the ROCK 5C, the CPU calling the search was pinned to either the high-performance cores (A76) or the low-power cores (A55), and the acquisition time was measured for each.

\begin{figure}[t]
\centering
\includegraphics[width=\linewidth]{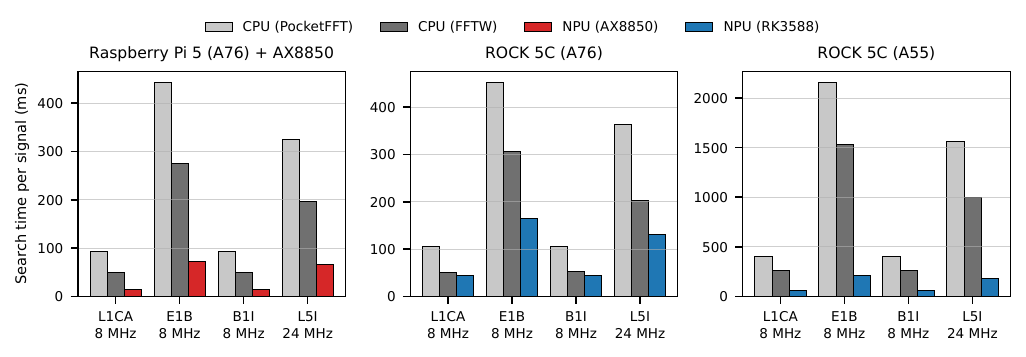}
\caption{Comparison of the search time per signal}\label{fig:acqtime}
\end{figure}

\begin{figure}[t]
\centering
\includegraphics[width=0.75\linewidth]{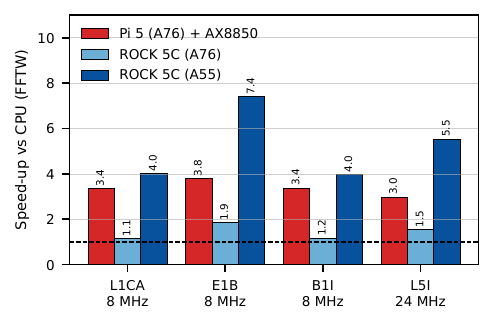}
\caption{Speed-up of the NPU relative to the CPU (FFTW) in search time. Mean of five trials. The lower the CPU performance (A55), the larger the speed-up}\label{fig:speedup}
\end{figure}

Table~\ref{tab:acq} and Fig.~\ref{fig:acqtime} show the search time, and Fig.~\ref{fig:speedup} shows the speed-up relative to the optimized CPU acquisition (using FFTW). The number of detected signals and their code phases were identical for all methods and all five trials.

\begin{table}[t]
\caption{Search time of acquisition (ms per signal, all PRNs, mean of five trials)}\label{tab:acq}
\centering
\begin{tabular}{@{}llrrr@{}}
\toprule
Platform & Signal & CPU PocketFFT & CPU FFTW & NPU \\
\midrule
\multirow{4}{*}{Pi 5 (A76) + AX8850} & L1CA 8 MHz & 94 & 51 & 15 \\
 & E1B 8 MHz & 444 & 275 & 72 \\
 & B1I 8 MHz & 94 & 51 & 15 \\
 & L5I 24 MHz & 326 & 197 & 66 \\
\midrule
\multirow{4}{*}{ROCK 5C (A76)} & L1CA 8 MHz & 106 & 52 & 45 \\
 & E1B 8 MHz & 453 & 306 & 165 \\
 & B1I 8 MHz & 106 & 52 & 45 \\
 & L5I 24 MHz & 364 & 203 & 132 \\
\midrule
\multirow{4}{*}{ROCK 5C (A55)} & L1CA 8 MHz & 400 & 263 & 65 \\
 & E1B 8 MHz & 2161 & 1536 & 207 \\
 & B1I 8 MHz & 400 & 259 & 65 \\
 & L5I 24 MHz & 1560 & 1002 & 181 \\
\bottomrule
\end{tabular}
\end{table}

On the Raspberry Pi~5 (A76), the search time with the AX8850 was 1/3.0--1/3.8 of that of the CPU (FFTW) and 1/4.9--1/6.3 of that of the CPU (PocketFFT). On the ROCK 5C, the NPU was most effective when the search was called on the A55, reducing the search time to 1/4.0--1/7.4 of that of the CPU (FFTW). Because the search time of the NPU differs little between the A76 and the A55, this confirms that the benefit of the NPU increases as the CPU performance decreases. Whereas the AX8850 achieved a speed-up of 3.4 times for L1 C/A, the speed-up for L1 C/A and B1I on the A76 of the ROCK 5C was only 1.1--1.2 times, because for the small search with $M = 16000$ the CPU (FFTW) is already fast and the inference time of the NPU is relatively large. For E1B, with a longer code and a larger DFT, the speed-up reached 1.9 times even on the A76.

\begin{figure}[t]
\centering
\includegraphics[width=\linewidth]{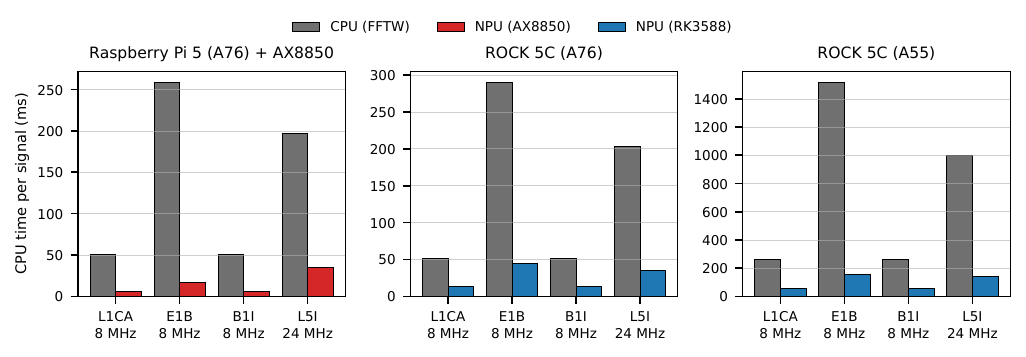}
\caption{CPU time used for acquisition. With the NPU, the CPU time is reduced to 1/5.6--1/15 on the Pi~5 and to 1/3.7--1/10 on the ROCK 5C}\label{fig:acqcpu}
\end{figure}

Figure~\ref{fig:acqcpu} compares the CPU time, i.e., the time the CPU spends on acquisition. With the NPU, the CPU time of acquisition was reduced to 1/5.6--1/15 on the Raspberry Pi~5 and to 1/3.7--1/10 on the ROCK 5C. On the Raspberry Pi~5, the CPU performs the carrier wipe-off of the IF data, their DFT by FFTW, and the conversion to integers, but because the non-coherent integration and the peak detection are performed on the NPU, almost no processing of the outputs remains on the CPU. For L5I (24~MHz) with a large DFT, the share of the FFT on the CPU is large, and the reduction was limited to 1/5.6. The NPU reduced the CPU time for all models, and the CPU time saved in acquisition can be used for tracking.

\subsection{Acquisition Sensitivity}

The acquisition sensitivity with the CPU and the NPU was evaluated using synthetic IF data. In the same format as the FE 4CH (24~MHz, 2-bit IQ), PRN~1 signals of L1 C/A, E1B, and L5I were generated without navigation data, with random code phases, Doppler frequencies ($\pm$4~kHz, including code Doppler), and carrier phases, and each was searched 100 times at each $C/N_0$ from 30 to 45~dB-Hz in 1-dB steps. A detection within $\pm$1.5 samples in code phase and $\pm$1 bin in Doppler was counted as correct. False detections were also examined with noise-only data.

\begin{figure}[t]
\centering
\includegraphics[width=\linewidth]{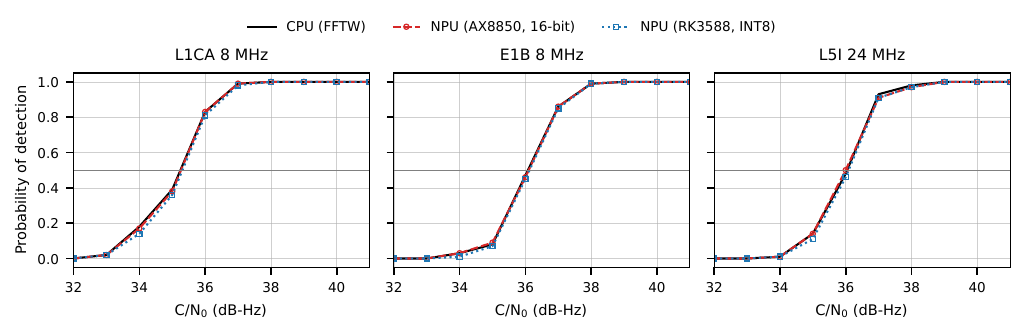}
\caption{Detection probability of acquisition versus $C/N_0$. Synthetic IF data, 100 trials at each $C/N_0$. The $C/N_0$ at a detection probability of 0.5 differs between the NPU and the CPU by 0.00--0.06~dB. No false detection occurred in the noise-only trials for any method}\label{fig:sens}
\end{figure}

Figure~\ref{fig:sens} shows the acquisition sensitivity. The $C/N_0$ at a detection probability of 0.5, estimated by linear interpolation, was 35.25~dB-Hz for L1 C/A, 36.08~dB-Hz for E1B, and 36.04~dB-Hz for L5I with the CPU (FFTW), and the differences from the NPU were 0.00--0.06~dB. The differences in detection probability at each $C/N_0$ were 0.06 or less, comparable to the standard error of 100 trials near a detection probability of 0.5 (about 0.05). The differences in the estimated $C/N_0$ were also within 0.1~dB. With noise-only data, no false detection occurred in 100 trials for any method. Within these evaluation conditions and numbers of trials, no clear sensitivity loss due to the quantization of the NPU (INT8 on the RK3588 and 16-bit on the AX8850) was observed.

\section{Evaluation 3: Direct Acquisition of Long Codes}\label{sec:eval2}

\subsection{Method}

For long codes such as GPS L1CP (10~ms) and L2CM (20~ms), the number of code phases to search grows with the code length, and the number of Doppler bins also grows because the bin spacing becomes narrower. Ordinary receivers therefore perform assisted acquisition over about three bins, using the time and Doppler obtained from a signal acquired first, such as L1 C/A~\cite{VanDiggelen2009AGPS}. However, direct acquisition of long-code signals would allow them to be acquired without relying on other signals.

We therefore searched GPS L1CP (10~ms, $M = 160000$, 201 bins) and L2CM (20~ms, $M = 320000$, 401 bins) directly over the full range of $\pm$5~kHz without assistance and evaluated the search time and CPU time. Both L1CP and L2CM were searched with the NPU on both SBCs.

\subsection{Results}

\begin{figure}[t]
\centering
\includegraphics[width=\linewidth]{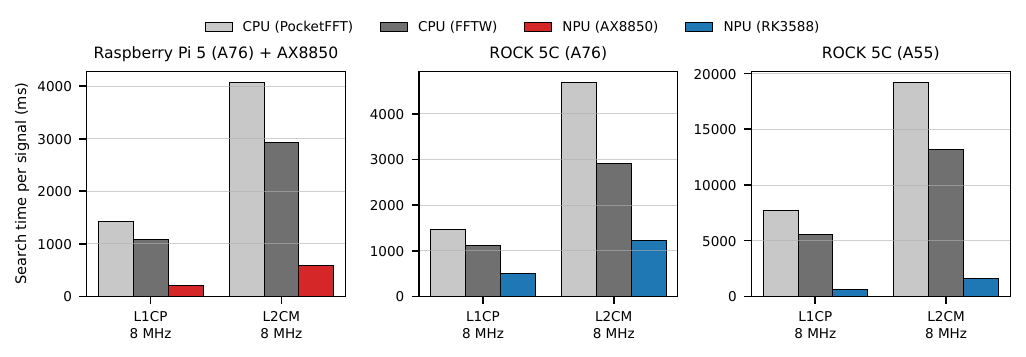}
\caption{Comparison of the search time of the direct search of long codes}\label{fig:long}
\end{figure}

\begin{figure}[t]
\centering
\includegraphics[width=\linewidth]{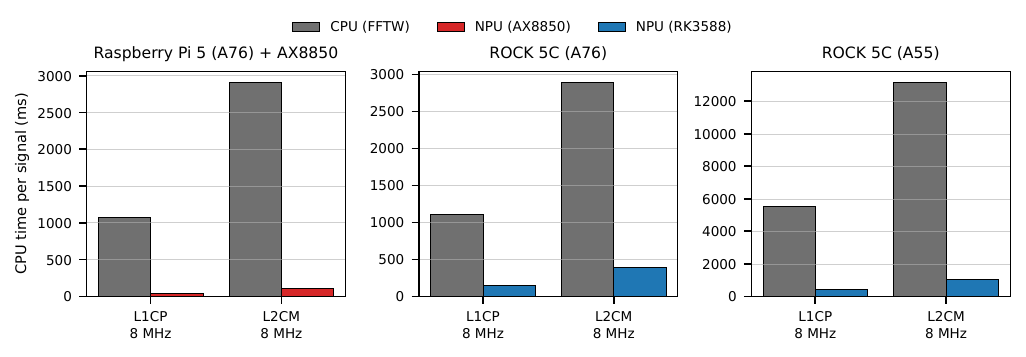}
\caption{Comparison of the CPU time in the direct search of long codes. With the NPU, the CPU time is reduced to about 1/26 on the Pi~5 and to 1/7--1/13 on the ROCK 5C}\label{fig:longcpu}
\end{figure}

Table~\ref{tab:long} and Fig.~\ref{fig:long} show the search times for the L1CP and L2CM signals, and Fig.~\ref{fig:longcpu} shows the CPU time required for acquisition. The number of detected signals and their code phases were identical for all methods and all five trials.

\begin{table}[t]
\caption{Direct search time of long codes (ms per signal, mean of five trials)}\label{tab:long}
\centering
\begin{tabular}{@{}llrrr@{}}
\toprule
Platform & Signal & CPU PocketFFT & CPU FFTW & NPU \\
\midrule
\multirow{2}{*}{Pi 5 (A76) + AX8850} & L1CP 8 MHz & 1426 & 1088 & 206 \\
 & L2CM 8 MHz & 4082 & 2936 & 581 \\
\midrule
\multirow{2}{*}{ROCK 5C (A76)} & L1CP 8 MHz & 1473 & 1123 & 513 \\
 & L2CM 8 MHz & 4696 & 2918 & 1231 \\
\midrule
\multirow{2}{*}{ROCK 5C (A55)} & L1CP 8 MHz & 7748 & 5557 & 645 \\
 & L2CM 8 MHz & 19241 & 13213 & 1584 \\
\bottomrule
\end{tabular}
\end{table}

As these results show, the speed-up of the NPU was larger for long codes than for short codes. Compared with the CPU (FFTW), NPU-based acquisition was 5.3 times faster for L1CP and 5.1 times faster for L2CM on the Raspberry Pi~5, 2.2--2.4 times faster on the A76 of the ROCK 5C, and 8.3--8.6 times faster on the A55. Furthermore, the CPU time used for acquisition was reduced to about 1/26 on the Raspberry Pi~5 and to 1/7--1/13 on the ROCK 5C. We attribute this to the fact that the larger the DFT, the more efficient the matrix operations on the NPU and the smaller the share of the processing on the CPU.

Based on the mean search times in Table~\ref{tab:long}, a direct search of all 32 PRNs of L2CM with the CPU (FFTW) on the A55 takes about 423~s, whereas it takes about 51~s with the NPU, making the direct acquisition of long codes practical. On the Raspberry Pi~5, the direct search of all 32 PRNs of L2CM takes about 94~s with the CPU (FFTW) and about 19~s with the NPU. These results confirm that the NPU enables the direct acquisition of long codes within a practical time.

\section{Evaluation 4: End-to-End Evaluation of the Receiver}\label{sec:eval3}

\subsection{Method}

The effect of faster acquisition on the whole receiver was evaluated with an end-to-end test in which the receiver tracks a fixed set of signals. First, N signals were selected as the fixed set from the 157 signals (25 signal types) in the 300-s IF data that were tracked until the end. If N exceeds 157, the same signals are tracked more than once. In addition, 142 signals that were never found in the recording were added as channels that keep searching, which emulates the load of continuously searching for signals of invisible satellites in real environments. The IF data were fed to the receiver in real time, and new searches were suspended while the usage of the IF data input buffer of the tracking exceeded 20\%. The NPU and the CPU (FFTW) were run alternately, five times each, and the numbers of acquired signals, tracking locks, and other metrics were evaluated.

The Pocket SDR receiver performs one search at a time, and each channel thread runs with a cycle of 50~ms. The time of a search is therefore quantized to multiples of 50~ms and is at least about 100~ms for short codes, so the speed of the initial acquisition depends on whether the computation of a search fits into this cycle. On the ROCK 5C, we therefore evaluated two conditions in which the search was pinned either to the high-performance cores (Cortex-A76) or to the low-power cores (Cortex-A55). The Raspberry Pi~5 has only Cortex-A76 cores, so no cores were pinned.

The evaluation metrics are the time evolution of the cumulative number of channels in the fixed set that have reached their first lock (hereafter, the number of locks), the time until 90\% of the fixed set reach their first lock ($t_{90}$), and the number of searches in 300~s. The Raspberry Pi~5 was tested with N = 140--240 and the ROCK 5C with N = 220--260, and we present the results for N = 175 on the Raspberry Pi~5 and N = 240 on the ROCK 5C. These are the largest values of N for which the IF buffer of the tracking did not overflow in any of the five trials with CPU-based acquisition. In addition, the upper limit of N that can be tracked in real time was evaluated from whether the IF buffer of the tracking overflowed.

\subsection{Results}

\begin{figure}[t]
\centering
\includegraphics[width=\linewidth]{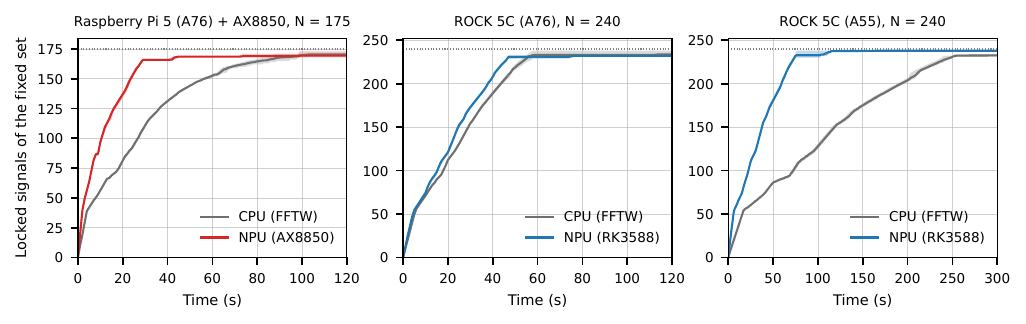}
\caption{Cumulative number of channels in the fixed set that have reached their first lock. N = 175 on the Pi~5 and N = 240 on the ROCK 5C; lines show the means and bands the ranges of five trials. On the ROCK 5C, the search is pinned to the A76 or the A55}\label{fig:lock}
\end{figure}

Table~\ref{tab:e2e} summarizes the end-to-end results, and Fig.~\ref{fig:lock} shows the time evolution of the number of locks. The NPU accelerated the initial acquisition, and the effect depended strongly on the performance of the cores that run the search. The number of channels locked within 30~s increased from 111 to 166 (1.50$\times$) on the Raspberry Pi~5, from 154 to 173 (1.12$\times$) on the high-performance cores of the ROCK 5C, and from 65 to 121 (1.86$\times$) on its low-power cores. $t_{90}$ was shortened from 66 to 26~s on the Raspberry Pi~5, from 49 to 42~s on the high-performance cores of the ROCK 5C, and from 215 to 66~s on its low-power cores. On the low-power cores, the CPU search is about five times slower than on the high-performance cores, which delays the initial acquisition considerably, whereas the search time of the NPU hardly changes on the low-power cores, so the difference becomes large. On the high-performance cores of the ROCK 5C, in contrast, both the CPU search and the NPU search fit into the search cycle of 100~ms, so the difference is small for short codes and arises mainly from the differences in the search times of E1B and the long codes.

On the ROCK 5C, the upper limit of the number of tracking channels that can be processed in real time was about 240 and hardly differed between the NPU and the CPU. On the RK3588, the CPU time of acquisition is reduced to 1/3.7--1/10 (Fig.~\ref{fig:acqcpu}), but the saved CPU time is spent on the additional searches and on tracking the signals locked earlier, and the limit is determined by the load of the tracking correlators. On the Raspberry Pi~5, in contrast, with CPU-based acquisition the IF buffer of the tracking overflowed in all five trials for N of 190 or more, and the limit was about 175 channels. With the AX8850 NPU, the buffer did not overflow in any of the five trials up to N = 230, and the limit increased to about 230 channels. This is because the AX8850 almost eliminates the CPU time of acquisition, and the saved CPU time can be used for tracking.

\begin{table}[t]
\caption{End-to-end test with the fixed set (300~s, mean of five trials)}\label{tab:e2e}
\centering
\begin{tabular}{@{}lllcc@{}}
\toprule
Platform (search cores) & N & Method & Locks at 10/30/60/300 s & $t_{90}$ (s) \\
\midrule
\multirow{2}{*}{Pi 5 (A76) + AX8850} & \multirow{2}{*}{175} & CPU & 56 / 111 / 153 / 170 & 66 \\
 & & NPU (AX8850) & 97 / 166 / 169 / 173 & 26 \\
\midrule
\multirow{2}{*}{ROCK 5C (A76)} & \multirow{2}{*}{240} & CPU & 71 / 154 / 233 / 237 & 49 \\
 & & NPU (RK3588) & 74 / 173 / 231 / 238 & 42 \\
\midrule
\multirow{2}{*}{ROCK 5C (A55)} & \multirow{2}{*}{240} & CPU & 32 / 65 / 91 / 233 & 215 \\
 & & NPU (RK3588) & 62 / 121 / 202 / 238 & 66 \\
\bottomrule
\end{tabular}
\end{table}

\begin{figure}[t]
\centering
\includegraphics[width=\linewidth]{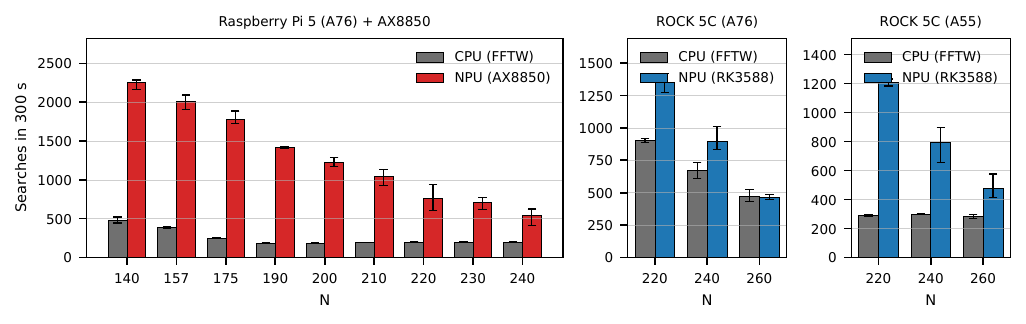}
\caption{Number of searches in 300~s for different N (mean and range of five trials)}\label{fig:search}
\end{figure}

Next, Table~\ref{tab:search} and Fig.~\ref{fig:search} show the number of searches of each method in the 300-s end-to-end test for different sizes N of the fixed set. As shown in Table~\ref{tab:search}, the number of searches in 300~s increased from a mean of 249 with the CPU to a mean of 1774 with the NPU (about 7.1 times) for N = 175 on the Raspberry Pi~5, from 675 to 899 (about 1.3 times) for N = 240 on the high-performance cores of the ROCK 5C, and from 300 to 794 (about 2.6 times) for N = 240 on its low-power cores. As shown in Fig.~\ref{fig:search}, on the ROCK 5C the difference becomes larger as the number of tracked signals decreases, whereas with many tracked signals and a heavy load, new searches are suspended and the difference becomes small. On the Raspberry Pi~5 with the AX8850, the NPU performed more searches for all N; with the CPU, the number of searches levelled off at about 190 for N of 190 or more, whereas with the NPU the receiver continued searching even under heavier loads. The increased number of searches leads to the faster initial acquisition (the larger number of locks at 30~s). In real environments where invisible signals are searched for continuously, more searches also mean faster reacquisition after blockage.

\begin{table}[t]
\caption{Number of searches (300~s, mean and range of five trials)}\label{tab:search}
\centering
\begin{tabular}{@{}llcc@{}}
\toprule
Platform (search cores) & N & CPU & NPU \\
\midrule
Pi 5 (A76) + AX8850 & 175 & 249 (243--254) & 1774 (1730--1885) \\
ROCK 5C (A76) & 240 & 675 (611--733) & 899 (836--1011) \\
ROCK 5C (A55) & 240 & 300 (298--301) & 794 (658--898) \\
\bottomrule
\end{tabular}
\end{table}

These results show that, in the end-to-end evaluation of the receiver, using the NPU for acquisition increases the number of searches, and that reducing the CPU load of acquisition raises the upper limit of the number of tracking channels.

\section{Discussion}\label{sec:disc}

\textbf{Conditions under which the NPU is effective}: The benefit of the NPU is larger for lower CPU performance and for larger DFTs (longer codes and higher sampling frequencies). Because the search time of NPU-based acquisition depends little on the performance of the CPU that calls the search, even a low-power core of an SBC can search almost as fast as a high-performance core. Moreover, when the NPU reduces the CPU time spent on acquisition, the freed CPU time can be used for other receiver processing. Indeed, with the AX8850, which almost eliminates the CPU time of acquisition, the number of tracking channels that can be processed in real time itself increased from about 175 to about 230. Offloading acquisition to the NPU thus not only accelerates acquisition but also increases the tracking capacity of the receiver by letting the CPU concentrate on tracking.

\textbf{NPU configuration and division of work}: The performance of the NPU is determined not only by computation but also by the passing of data to and from the host. For an NPU that shares memory with the CPU, the cost of passing data is almost zero and the computation on the NPU is the bottleneck; computing the forward DFT also on the NPU to minimize the CPU time was therefore suitable. For an NPU with its own memory, the amount and number of transfers can be the bottleneck; computing the DFT of the IF data only once on the CPU, sending it as integers, and performing the non-coherent integration and the peak detection on the NPU to minimize the transfers was suitable.

\textbf{Quantization precision and interference}: In our evaluation, the differences in the interpolated sensitivity between the quantized NPU and the CPU were within 0.06~dB. However, because the quantization scales are calibrated with interference-free signals, strong narrowband or pulsed interference may concentrate the energy after the DFT into a few elements and saturate the quantization range~\cite{Balaei2008QuantCW}. Countermeasures against interference are left for future work and are outside the scope of this evaluation.

\textbf{Application to other NPUs}: The proposed method expresses the parallel code search using only basic operations supported by many NPUs: $1\times1$ convolution, element-wise product, and transpose. By changing $M$, $K_c$, $A \times B$, and the number of correlation segments in the model generation scripts, and by choosing the division of work between the CPU and the NPU according to the configuration of the NPU, it can be applied to other NPUs and sampling frequencies.

\section{Conclusions}\label{sec:concl}

This paper proposed using the NPU of an SBC as a dedicated processor for GNSS signal acquisition. The DFT of the parallel code search was expressed, by the four-step DFT, as matrix products of small DFTs with $1\times1$ convolutions and transposes, and the quantized models were implemented as a plugin of Pocket SDR. On the NPU, this DFT by matrix products was shown to be about 3.5 times faster than the ordinary FFT, which requires about 1/8 of the operations. Furthermore, according to the configuration of the NPU, we showed which parts of the acquisition should be performed on the CPU and which on the NPU. Evaluations with the NPUs of a Raspberry Pi~5 with an AX8850 and of an RK3588 showed that, compared with an optimized CPU FFT, the acquisition of short codes was 3.0--3.8 times faster on the Raspberry Pi~5 and 4.0--7.4 times faster on the A55 of the RK3588. The direct acquisition of long codes over the full Doppler range was 5.1--5.3 times faster on the Raspberry Pi~5 and 8.3--8.6 times faster on the A55, making acquisition without assistance practical. In the whole receiver, the number of channels locked within 30~s increased, and on the Raspberry Pi~5 with the AX8850, the CPU load of acquisition was almost eliminated and the number of tracking channels that can be processed in real time increased from about 175 to about 230. These results show that using the NPU on SBCs with limited CPU resources is effective for achieving both GNSS signal acquisition and tracking.

\section*{List of abbreviations}

ADC: analog-to-digital converter; ASIC: application-specific integrated circuit; CPU: central processing unit; DFT: discrete Fourier transform; FFT: fast Fourier transform; FIR: finite impulse response; FPGA: field-programmable gate array; GNSS: global navigation satellite system; GPU: graphics processing unit; IDFT: inverse discrete Fourier transform; IF: intermediate frequency; IFFT: inverse fast Fourier transform; NPU: neural processing unit; PCIe: Peripheral Component Interconnect Express; PRN: pseudorandom noise; RMS: root mean square; SBC: single-board computer; SDR: software-defined radio; SIMD: single instruction, multiple data; SoC: system on a chip; TOPS: tera operations per second; TPU: tensor processing unit.

\ifanon\else
\section*{Declarations}

\subsubsection*{Availability of data and materials}
The source code of PocketSDR-NPU, the model generation scripts, the generated models, the evaluation scripts, and a download link to the IF data used in this study are available in the PocketSDR-NPU repository on GitHub (\anon{URL withheld for anonymous review}{\url{https://github.com/taroz/PocketSDR-NPU}}).
Project name: PocketSDR-NPU;
operating system: Linux (Raspberry Pi OS/Debian, Radxa Debian);
programming languages: C and Python;
other requirements: AXCL runtime for the AX8850 and RKNN runtime for the RK3588;
license: BSD 2-clause license (the same as Pocket SDR).

\subsubsection*{Competing interests}
The \anon{author(s) declare}{author declares} that there are no competing interests.

\subsubsection*{Funding}
Not applicable.

\subsubsection*{Authors' contributions}
\anon{Withheld for anonymous review.}{TS conceived the study, developed the software, performed the experiments, analyzed the results, and wrote the manuscript. The author read and approved the final manuscript.}

\subsubsection*{Acknowledgements}
\anon{Withheld for anonymous review.}{The author is deeply grateful to Mr. Tomoji Takasu for developing Pocket SDR, on which this work is based, and for making it publicly available as open source.}
\fi

\bibliographystyle{splncs04}
\bibliography{refs}

\end{document}